\documentclass{article}

\usepackage[numbers]{natbib}

    \usepackage[final]{timeseries_workshop}

\usepackage[utf8]{inputenc} 
\usepackage[T1]{fontenc}    
\usepackage{hyperref}       
\usepackage{url}            
\usepackage{booktabs}       
\usepackage{amsfonts}       
\usepackage{nicefrac}       
\usepackage{microtype}      
\usepackage{xcolor}         
\usepackage{amsmath}

\hypersetup{
    colorlinks=true,
    urlcolor=blue
}
\usepackage{graphicx}

\title{Fine-Tuning a Time Series Foundation Model with Wasserstein Loss}

\author{%
  Andrei Chernov \\
  Independent Researcher \\
  \texttt{chernov.andrey.998@gmail.com} \\
}

\begin{document}

\maketitle

\begin{abstract}
Inspired by recent advancements in large language models (LLMs) for Natural Language Processing (NLP), there has been a surge in research focused on developing foundational models for time series forecasting. One approach involves training LLM architectures on tokenized time series data using cross-entropy loss. Although this method has demonstrated promising results, cross-entropy loss is primarily designed for classification tasks and does not account for the distance between classes. To address this limitation, we propose using the Wasserstein loss for such architectures. To validate our approach, we fine-tuned a foundational time series model on $22$ zero-shot datasets, comparing the performance of cross-entropy loss with that of Wasserstein loss. Our results demonstrate that replacing cross-entropy loss with Wasserstein loss significantly improves point estimation.
\end{abstract}

\section{Introduction}
Time series forecasting is a well-known problem across various domains, such as finance, retail, and healthcare. Traditionally, it has been addressed using statistical models like ARIMA \citep{contreras2003arima} or Bayesian time series frameworks, such as Prophet \citep{taylor2018forecasting}. More recent approaches have applied deep learning models \citep{DBLP:journals/corr/abs-2004-10240, DBLP:journals/corr/FlunkertSG17}, which have demonstrated promising results in several competitions, such as the M5 competition \citep{MAKRIDAKIS20221346}.

At the same time, we are witnessing significant progress in foundational large language models (LLMs) for natural language processing (NLP) tasks \citep{radford2019language, team2023gemini}. This raises the question of whether massive pretrained deep learning models can also perform well on time series data. However, there is a clear structural difference: NLP data consists of text, which can be tokenized, with each token treated as a class, naturally framing the problem as a classification task. This approach typically uses cross-entropy as the loss function, which treats all errors equally. If the model predicts the wrong class, the penalty remains the same regardless of which incorrect class is chosen. In contrast, time series data typically represents a continuous domain, leading to a regression problem, which motivates the use of distance-based loss functions, such as mean squared error. This distinction makes it challenging to directly apply LLM architectures to the time series domain.

A first step to overcoming this distinction is to tokenize time series values to create a fixed vocabulary. This allows each token to be treated as a class, enabling the use of cross-entropy loss, as demonstrated in \citep{ansari2024chronos}. Although this approach has shown significant performance improvements, it still ignores the distances between classes. In this paper, we extend this approach by proposing to replace cross-entropy loss with Wasserstein loss, which accounts for the distance between classes. To validate our idea, we fine-tuned one of the models from \citep{ansari2024chronos} using both cross-entropy loss and Wasserstein loss on zero-shot datasets, i.e., datasets the model had not seen during training. We chose not to train models from scratch with Wasserstein loss due to: (a) the high cost of training from scratch, and (b) the fact that foundational time series models are still significantly smaller compared to LLMs, making fine-tuning in the time series domain more efficient and desirable for industrial applications.

The rest of the paper is organized as follows: Section \ref{background} provides an overview of time series forecasting and Wasserstein loss. In Section \ref{main}, we present our approach for incorporating topology into tokenized deep time series forecasting through the application of Wasserstein loss. Section \ref{experiments} details the zero-shot datasets used for fine-tuning, the evaluation metrics, and our results. Finally, Section \ref{discussion} concludes the paper and explores potential directions for future research.

\section{Background}
\label{background}

\paragraph{Time Series Forecasting.} The time series forecasting problem can be formulated as follows: given a time series dataset $x_1, x_2, \dots, x_n$, the goal is to find the distribution $p(x_{n+1}, x_{n+2}, \dots, x_{n+k} | x_1, x_2, \dots, x_n)$, where all $x_i \in \mathbb{R}$, and $k$ represents the forecast horizon, referring to the number of future steps the model needs to predict. It is common to use an autoregressive approach, where one step is forecast at a time, and the result is appended to the input sequence to predict the next value. Another common simplification is to limit the model’s input to only the last $m$ values of the time series. These two modifications simplify the original task to: $p(x_{n+1} | x_{n-m+1}, x_{n-m+2}, \dots, x_n)$, where $m$ represents the context length. 

There are two types of models for time series forecasting: local and global. Traditional statistical models, such as ARIMA \citep{contreras2003arima}, fit a separate model for each time series. These models are considered local because a trained model can forecast only one specific time series. In contrast, deep learning approaches train a model on a dataset containing multiple time series, allowing a single model to forecast across a set of time series \citep{DBLP:journals/corr/abs-2004-10240, DBLP:journals/corr/FlunkertSG17}. However, these models are typically effective only for a limited number of time series. Recent research \citep{ansari2024chronos, das2024decoderonlyfoundationmodeltimeseries} in foundational time series models aims to build models that can achieve reasonable accuracy across a wide range of datasets.

\paragraph{Wasserstein Loss.} The Wasserstein metric\footnote{In this paper, we use the terms Wasserstein loss, distance, and metric interchangeably.} is widely utilized in the field of Optimal Transport \citep{peyré2020computationaloptimaltransport} as a tool for calculating distances between distributions. One of the key advantages of Wasserstein loss, which we leverage in this paper, is that it takes into account the underlying geometry of the space. Consider a simple example: suppose we have three uniformly distributed univariate random variables: $X_1 \sim U[0,1]$, $X_2 \sim U[1,2]$, and $X_3 \sim U[10,11]$. In this case, the Wasserstein distance has a closed-form solution and is given by $\frac{1}{2}|E(X_i) - E(X_j)|$, where $1 \leq i, j \leq 3$. Thus, the Wasserstein distance between $X_1$ and $X_2$ is $0.5$, and between $X_1$ and $X_3$ it is $5$, reflecting the difference between the domains of the random variables. In the general case, the Wasserstein distance between two distributions $P$ and $Q$ can be defined as follows:

\begin{equation}
W_p(P, Q) = \inf_{\gamma \in \Gamma(P, Q)} \left(\mathbb{E}_{(x,y) \sim \gamma} [D(x, y)^p] \right)^{1/p}
\label{eq:wasserstein_distance}
\end{equation}

where $\gamma \in \Gamma(P, Q)$ denotes the set of all possible joint distributions $\gamma (x,y)$ whose marginals are equal to $P$ and $Q$, respectively. One prominent application of the Wasserstein distance in deep learning is in Wasserstein Generative Adversarial Networks \citep{arjovsky2017wassersteingan}. A key challenge with Wasserstein loss is that, in most cases, it is computationally expensive \citep{DBLP:journals/corr/FrognerZMAP15}. As we will discuss in Section \ref{main}, in our case, the Wasserstein loss has a closed-form solution that avoids the need for intensive computations.

\section{Wasserstein Deep Learning Model for Tokenized Time Series}
\label{main}
In this section, we discuss our approach to applying Wasserstein loss to a tokenized deep learning model. Our method can be generalized to tasks where the relative distance between predicted classes is crucial.

\subsection{Time Series Preprocessing}
In this paper, we fine-tuned a model from \citep{ansari2024chronos}, consequently we apply the same mean absolute scaling \citep{salinas2019deeparprobabilisticforecastingautoregressive} and quantization algorithm. Mean absolute scaling normalizes the data by the absolute mean, defined as $s = \frac{1}{n} \sum_{i=1}^n |x_i| $, which is calculated on the training data. The scaled data is given by $y_i = \frac{x_i}{s}$. To perform quantization, we first need to set the minimum value ($y_{min}$) and the maximum value ($y_{max}$) for the mean-scaled time series. The quantization process constructs a uniform grid from $y_{min}$ to $y_{max}$. The centroids of the grid cells are defined as $c_i = y_{min} + (i-1)\cdot\frac{y_{max} - y_{min}}{d-1}$, where $i \in {1, 2, \dots, d}$. Denoting the boundaries between cells $b_i = \frac{1}{2}(c_i + c_{i+1})$, where $i \in {1, 2, \dots, d-1}$ the tokenization function $q : \mathbb{R} \rightarrow {0, 1, \dots, d-1}$ is defined as $q(y) = \sum_{i=1}^{d-1} \mathbf{1}_{{y < b_i}}(y)$, where $\mathbf{1}_{A}(y)$ represents the indicator function.

\subsection{Model Architecture}
For the model architecture, we selected the pretrained Chronos-T5 (Small) model from \citep{ansari2024chronos}, which is based on the T5 architecture \citep{DBLP:journals/corr/abs-1910-10683}. The total number of tokens is $4096$, two of which are reserved for special symbols: \textit{PAD} and \textit{EOS}. The \textit{EOS} token denotes the end of the sequence, which is not necessary for time series applications, although its inclusion makes working with popular libraries more convenient. The \textit{PAD} token is used to align the number of samples in each time series during batch processing. Therefore, the number of grid cells is $d=4094$. The minimum value $y_{min}$ is set to $-15$, and the maximum value $y_{max}$ is set to $15$. Consequently, the distance between neighboring centroids is calculated as $r=\frac{y_{max} - y_{min}}{d-1} \approx 0.0073$.

\subsection{Loss Function}
In this paper, we primarily focus on point estimation for forecasting univariate values. Therefore, we model the target distribution as a degenerate random variable that takes only a single value. While this assumption is not strictly necessary, and any distribution could be assigned to the target variable, especially if the goal is to improve probabilistic forecasting, we advise against using overly complex distributions due to the potential computational intensity of calculating the Wasserstein loss. For the forecast distribution, we utilize the distribution over tokens, which is obtained from the neural network after the softmax operation. As a result, the Wasserstein distance needed to be calculated between a degenerate distribution and a discrete distribution. We define the distance between two tokens as the distance between their centroids: $D(y_i, y_j) = r \cdot |i - j|$. Thus, equation \ref{eq:wasserstein_distance} simplifies, and we obtain a closed-form formula for the Wasserstein metric:

\begin{equation}
W_p(Y_a, \hat{Y}) = r \cdot \left(\sum_{i=1}^{d} \alpha_i \cdot |i - a|^p \right)^{1/p}
\label{eq:wasserstein_distance_1D}
\end{equation}

where $Y_a$ represents the ground truth degenerate distribution, equal to the token $a$, $\hat{Y}$ is the forecasted distribution over tokens, $d$ is the number of tokens, and $\alpha_i$ is the predicted probability for token $i$. Note that if the model output were a scalar, this formula would reduce to well-known regression losses: the absolute error (AE) when \( p = 1 \) and the squared error (SE) when \( p = 2 \). The relationship between Wasserstein losses and AE/SE is explored further in Appendix \ref{appendix_losses}.

\subsection{Forecasting and Evaluation Metrics}

We maintain the same forecasting procedure and evaluation metrics as in \citep{ansari2024chronos} to ensure result comparability. We use autoregressive sampling from the predicted distribution over tokens. To convert a token back to the original time series format, we first apply the detokenization function, which returns the centroid of the bin, $q^{-1}(j) = c_{j+1}$, and then multiply the result by the scaling factor $s$.

For point estimation, we take the median forecast from the model and evaluate it using the mean absolute scaled error (MASE) \citep{hyndman2006another}. To assess the probabilistic forecast, we estimate the quantiles using $20$ sample forecast paths and apply the weighted quantile loss (WQL) on nine uniformly spaced quantile levels: ${0.1, 0.2, \dots, 0.9}$.

To aggregate scores across different datasets, we compute the relative score of each model by dividing the model’s score by the score of a seasonal naive forecast, then aggregate the relative scores across all datasets using the geometric mean to obtain the final metric.

\section{Experiments}
\label{experiments}

\subsection{Datasets}

For our experiments, we selected the zero-shot datasets from \citep{ansari2024chronos}, as these data were not seen by the model during training. To ensure reliable evaluation results, we filtered out datasets with fewer than $50$ time series, leaving $22$ datasets for experimentation. The last $k$ observations of each time series were allocated to the test set, while the remaining data were used for fine-tuning. The offset $k$ is unique to each dataset, and we maintained the same offsets as in \citep{ansari2024chronos}.

\subsection{Fine-Tuning Results}
As discussed in Section \ref{main}, we fine-tuned the pretrained Chronos-T5 (Small) model\footnote{The code is available at \href{https://github.com/ChernovAndrey/chronos-forecasting-wasserstein.git}{https://github.com/ChernovAndrey/chronos-forecasting-wasserstein.git}}. For each dataset, we conducted $1000$ fine-tuning steps, with the initial learning rate set to $0.001$, which linearly decreased to $0$ over the course of the steps. We fine-tuned the model using three different loss functions. The first two, Wasserstein-1 (W1) and Wasserstein-2 (W2), correspond to equation \ref{eq:wasserstein_distance_1D}, with $p=1$ and $p=2$, respectively. The third loss function is the standard cross-entropy loss. Additionally, we calculated metrics for the model without fine-tuning.

\begin{figure}[t]  
  \centering
  \begin{minipage}[c]{0.48\textwidth}
    \centering
    \includegraphics[width=\textwidth]{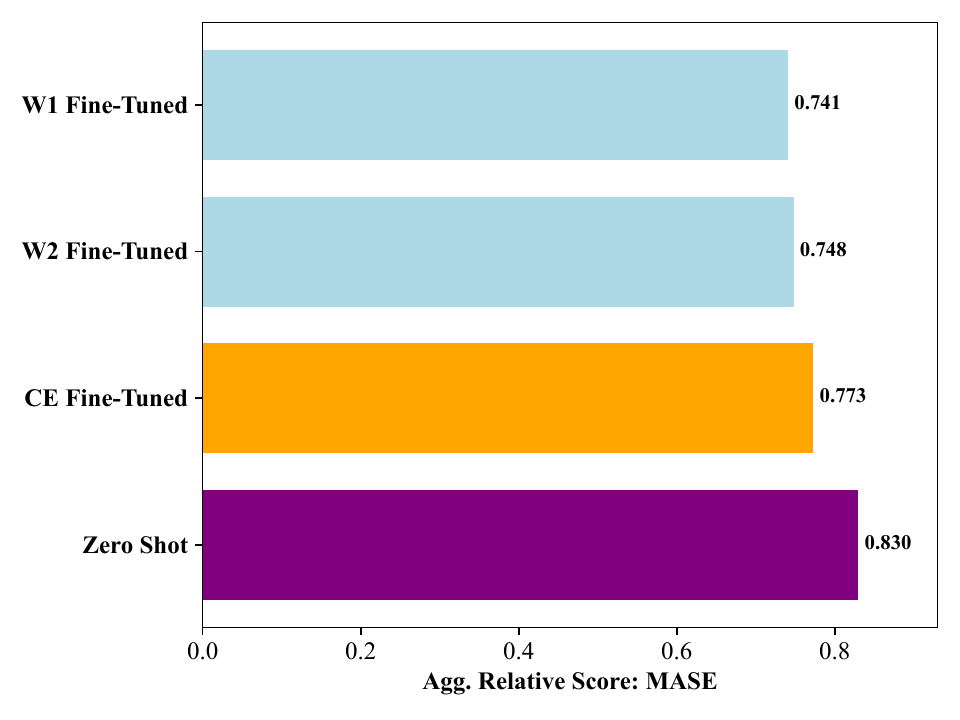}
    \caption{Performance Comparison: MASE}  
    \label{fig:mase}  
  \end{minipage}
  \hfill
  \begin{minipage}[c]{0.48\textwidth}
    \centering
    \includegraphics[width=\textwidth]{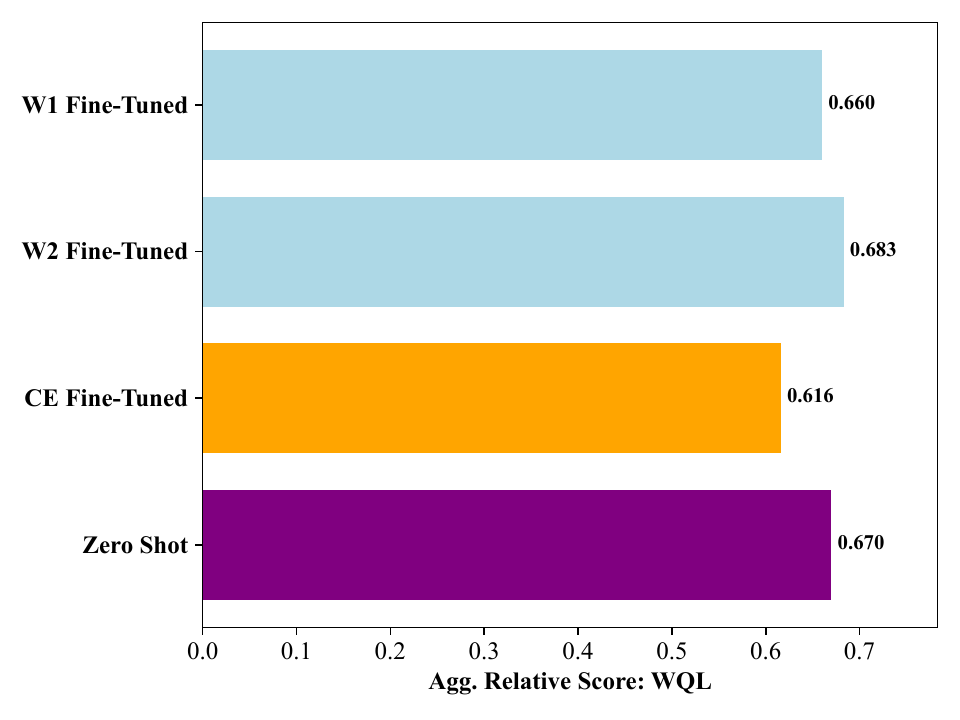}
    \caption{Performance Comparison: WQL}  
    \label{fig:wql}  
  \end{minipage}
\end{figure}

Figures \ref{fig:mase} and \ref{fig:wql} present the results for MASE and WQL, respectively. Appendix \ref{appendix} provides the MASE and WQL values for each dataset. The Wasserstein loss significantly outperforms cross-entropy loss in point estimation; however, we observe a degradation in the WQL metric. This is a direct result of the loss design. Since we use a degenerate distribution as the target in the Wasserstein loss, the forecasted distribution becomes sharper and less suitable for quantile estimation compared to cross-entropy loss. See Appendix \ref{appendix_dist} for futher details.

\section{Discussion}
\label{discussion}
In this paper, we proposed an approach to applying Wasserstein loss to large language model (LLM) architectures, originally designed for NLP tasks, to account for the topology of the space in domains where the distance between classes is important, particularly in the time series domain. To validate our approach, we demonstrated that fine-tuning the Chronos Small model with Wasserstein loss improves point estimation compared to fine-tuning with cross-entropy loss.

\paragraph{Future Work} A key area for future work is training a foundational time series model from scratch using Wasserstein loss. Although probabilistic forecasting was not the main focus of this work, the model’s ability to capture uncertainty is crucial and should be explored in future research.


\newpage

\bibliographystyle{plain}
\bibliography{main}

\medskip

\newpage

\appendix

\section{Results for each dataset}
\label{appendix}
Tables \ref{mase-table} and \ref{wql-table} provide a comparison of the MASE and WQL metrics, respectively, between fine-tuning with Wasserstein-1 (W1) loss and cross-entropy (CE) loss across $22$ datasets. Point estimation with W1 is worse than with CE in only $2$ of the datasets.

\begin{table}
  \caption{Comparison of MASE Scores}
  \label{mase-table}
  \centering
  \begin{tabular}{lccc}
    \toprule
    Dataset               & MASE CE   & MASE W1 & $\Delta$ \\
    \midrule
    monash\_m1\_yearly    & 3.819  & 3.291    & 0.528         \\
    monash\_m3\_yearly    & 3.310  & 2.902    & 0.408         \\
    monash\_tourism\_yearly & 3.051  & 2.762    & 0.289         \\
    monash\_m1\_quarterly & 1.790  & 1.658    & 0.132         \\
    monash\_tourism\_quarterly & 1.633  & 1.523    & 0.110         \\
    monash\_nn5\_weekly   & 0.978  & 0.875    & 0.103         \\
    monash\_tourism\_monthly & 1.525  & 1.424    & 0.101         \\
    m4\_yearly           & 3.087  & 3.008    & 0.079         \\
    monash\_fred\_md      & 0.615  & 0.557    & 0.058         \\
    monash\_cif\_2016     & 1.000  & 0.959    & 0.041         \\
    monash\_m3\_quarterly & 1.215  & 1.173    & 0.042         \\
    monash\_traffic      & 0.742  & 0.716    & 0.026         \\
    m5                    & 0.925  & 0.900    & 0.025         \\
    monash\_car\_parts    & 0.815  & 0.796    & 0.019         \\
    m4\_quarterly        & 1.134  & 1.123    & 0.011         \\
    monash\_weather      & 0.835  & 0.829    & 0.006         \\
    dominick             & 0.797  & 0.791    & 0.006         \\
    monash\_m1\_monthly   & 1.104  & 1.099    & 0.005         \\
    monash\_hospital     & 0.687  & 0.684    & 0.003         \\
    nn5                  & 0.553  & 0.553    & 0.000         \\
    monash\_m3\_monthly   & 0.841  & 0.913    & -0.072        \\
    monash\_covid\_deaths & 34.582 & 35.314   & -0.732        \\
    \bottomrule
  \end{tabular}
\end{table}

\begin{table}
  \caption{Comparison of WQL Scores}
  \label{wql-table}
  \centering
  \begin{tabular}{lccc}
    \toprule
    Dataset               & WQL CE    & WQL W1  & $\Delta$ \\
    \midrule
    monash\_tourism\_yearly & 0.161  & 0.110    & 0.051        \\
    monash\_m1\_quarterly & 0.098  & 0.068    & 0.030        \\
    monash\_m3\_yearly    & 0.178  & 0.170    & 0.008        \\
    monash\_cif\_2016     & 0.015  & 0.009    & 0.006        \\
    monash\_tourism\_quarterly & 0.064  & 0.065    & -0.001       \\
    monash\_nn5\_weekly   & 0.095  & 0.098    & -0.003       \\
    monash\_m3\_quarterly & 0.077  & 0.081    & -0.004       \\
    monash\_covid\_deaths & 0.039  & 0.043    & -0.004       \\
    monash\_m1\_monthly   & 0.152  & 0.156    & -0.004       \\
    monash\_traffic      & 0.235  & 0.241    & -0.006       \\
    monash\_tourism\_monthly & 0.083  & 0.092    & -0.009       \\
    m4\_quarterly        & 0.077  & 0.086    & -0.009       \\
    monash\_m1\_yearly    & 0.104  & 0.115    & -0.011       \\
    monash\_hospital     & 0.057  & 0.070    & -0.013       \\
    m4\_yearly           & 0.111  & 0.125    & -0.014       \\
    nn5                  & 0.152  & 0.170    & -0.018       \\
    monash\_m3\_monthly   & 0.094  & 0.114    & -0.020       \\
    monash\_weather      & 0.141  & 0.164    & -0.023       \\
    monash\_fred\_md      & 0.022  & 0.067    & -0.045       \\
    dominick             & 0.327  & 0.392    & -0.065       \\
    monash\_car\_parts    & 0.899  & 1.000    & -0.101       \\
    m5                   & 0.582  & 0.698    & -0.116       \\
    \bottomrule
  \end{tabular}
\end{table}

\section{Relationship between Wasserstein Loss and Common Regression Losses}
\label{appendix_losses}
In this section, we demonstrate that the absolute error (AE) and squared error (SE) of a forecast’s expected value and its target, defined as follows:
\[
\text{AE}(Y_a, \mathbb{E}[\hat{Y}]) = |\mathbb{E}[\hat{Y}] - a|, \quad \text{and} \quad \text{SE}(Y_a, \mathbb{E}[\hat{Y}]) = (\mathbb{E}[\hat{Y}] - a)^2,
\]
serve as lower bounds for the Wasserstein losses:

\begin{equation}
W_p(Y_a, \hat{Y}) = r \cdot \left(\sum_{i=1}^{d} \alpha_i \cdot |i - a|^p \right)^{1/p} = 
r \cdot \left(\mathbb{E}[|i - a|^p] \right)^{1/p} \geq r \cdot \left(|\mathbb{E}[Y] - a|^p\right)^{1/p},
\label{eq:mae_rmse}
\end{equation}

where \( p \in \{1, 2\} \). The last inequality follows from Jensen’s inequality, since the function \( f(i) = |i - a|^p \) is convex for \( p = 1 \) or \( p = 2 \).

Thus, we obtain:
\[
W_1(Y_a, \hat{Y}) \geq \text{AE}(Y_a, \mathbb{E}[\hat{Y}]) \quad \text{and} \quad W_2(Y_a, \hat{Y})^2 \geq \text{SE}(Y_a, \mathbb{E}[\hat{Y}]).
\]

Although we did not conduct experiments using AE and SE losses, exploring these could be a promising direction for future research.

\section{Distribution Forecasting}
\label{appendix_dist}
\begin{figure}[t]
  \centering
  \includegraphics[width=\textwidth]{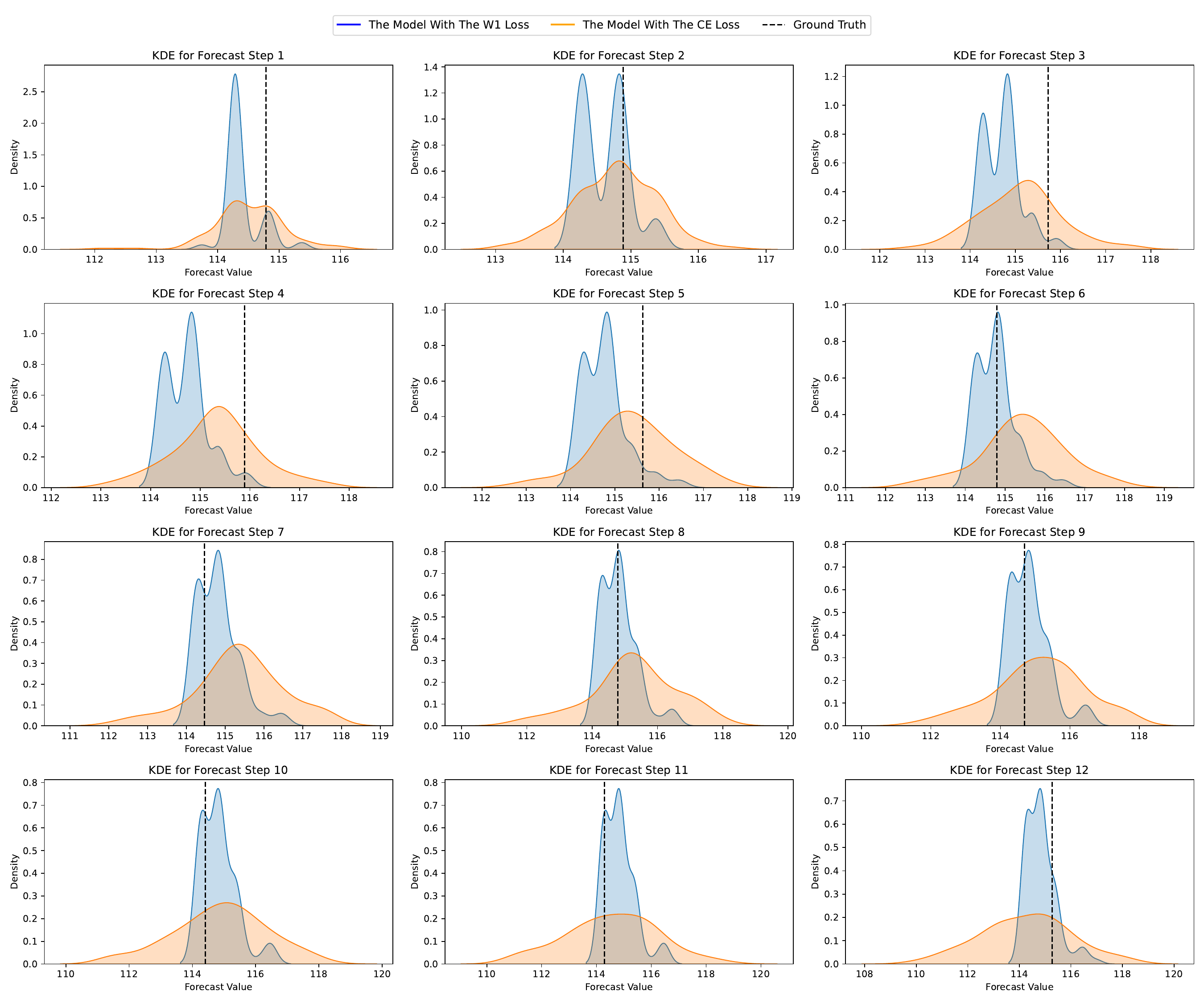}
  \caption{Kernel density estimation (KDE) comparison of forecasts on the FRED-MD dataset. The plot shows that the model trained with W1 loss produces significantly sharper forecast distributions compared to the model trained with cross-entropy loss.}
  \label{fig:comparison}
\end{figure}

In this section, we explore why models trained with the Wasserstein loss tend to exhibit worse WQL metrics compared to those trained with cross-entropy loss.

The asymptotic behavior of these loss functions diverges significantly as the number of tokens ($n$) increases. Consider a simple case where the predicted distribution consists of only two tokens with non-zero probabilities: $p$ and $1-p$, where $p \in (0, 1)$. Additionally, assume that the non-zero tokens are the first and last in the sequence, with the target corresponding to the token with probability $p$. In this scenario, the cross-entropy loss remains constant as $n$ increases, given by $-\log(p)$. In contrast, the Wasserstein loss depends on $n$ and diverges as $n$ approaches infinity, following $((1-p) \cdot (n-1)^p)^{1/p}$. This behavior resembles that of regression losses, resulting in a sharper predicted distribution, as illustrated in Figure \ref{fig:comparison}. Consequently, when the model’s predictions deviate significantly from the target, the quantile loss increases substantially.

We investigated these differences in forecasting using the Monash FRED-MD dataset, which contains 107 monthly time series representing various macroeconomic indicators from the Federal Reserve Bank. Our analysis showed that the model trained with the W1 loss performed better on 55 time series, while the model trained with the cross-entropy (CE) loss performed better on 52 time series. However, the aggregated WQL metric of the model trained with CE loss was significantly better. This is because, when the model’s predictions deviate notably from the ground truth, the quantile losses for the model trained with W1 loss are much higher. Nonetheless, since models trained with Wasserstein loss exhibit less bias in point estimation, incorporating sampling techniques like Monte Carlo Dropout could potentially enhance the quality of distribution forecasts.

\end{document}